\documentclass[pdflatex,sn-mathphys-num]{sn-jnl}

\usepackage{graphicx}%
\usepackage{multirow}%
\usepackage{amsmath,amssymb,amsfonts}%
\usepackage{amsthm}%
\usepackage{mathrsfs}%
\usepackage[title]{appendix}%
\usepackage{xcolor}%
\usepackage{textcomp}%
\usepackage{manyfoot}%
\usepackage{booktabs}%
\usepackage{algorithm}%
\usepackage{algorithmicx}%
\usepackage{algpseudocode}%
\usepackage{listings}%
\usepackage{geometry}

\theoremstyle{thmstyleone}%
\theoremstyle{thmstyletwo}%

\theoremstyle{thmstylethree}%
\usepackage{caption}
\begin{document}

\title[CogRad]{CogRad: A Cognitively-Inspired Multi-Agent Framework for Radiology Report Generation}

\raggedbottom
\author*[1,2]{\fnm{Saif Ur Rehman} \sur{Khan}}\email{saif\_ur\_rehman.khan@dfki.de}
\author[1]{\fnm{Hasaan} \sur{Maqsood}}\email{hasaankhattak159@gmail.com}

\author[1,2]{\fnm{Sebastian} \sur{Vollmer}}\email{sebastian.vollmer@dfki.de}

\author[1,2]{\fnm{Andreas} \sur{Dengel}}\email{andreas.dengel@dfki.de}

\author*[1,2,3,4]{\fnm{Muhammad Nabeel} \sur{Asim}}\email{muhammad\_nabeel.asim@dfki.de}

\affil[1]{\orgdiv{Department of Computer Science}, \orgname{Rhineland-Palatinate Technical University of
Kaiserslautern-Landau}, \orgaddress{\city{Kaiserslautern}, \postcode{67663}, \country{Germany}}}

\affil[2]{\orgname{German Research Center for Artificial Intelligence}, \orgaddress{\city{Kaiserslautern}, \postcode{67663}, \country{Germany}}}

\affil[3]{\orgname{Intelligentx GmbH (intelligentx.com)}, \orgaddress{\city{Kaiserslautern}, \country{Germany}}}

\affil[4]{\orgname{Department of Core Informatics, Graduate School of Informatics, Osaka Metropolitan University}, \orgaddress{\city{Sakai, 599-8531}, \country{Japan}}}
\abstract{Automated radiology report generation (RRG) can ease radiologist workload, yet most existing systems produce a report in a single forward pass, with no mechanism to check a claim against the image or revisit a finding once stated. We present CogRad, a cognitively inspired multi-agent framework that structures generation around four stages of a radiologist's reading process. A Scout agent discovers anatomical regions directly from image patches via slot attention and assigns region and disease-level triage scores; an Investigator agent concentrates representational capacity on the regions Scout flags as suspicious; a Writer agent compiles these signals into a disease gated visual prefix for a large language model; and a Verifier agent supervises training with a visual entailment loss and, at inference, re-examines its own draft sentence by sentence, regenerating any report it judges insufficiently grounded. On CheXpert Plus, CogRad attains a BLEU-4 of 0.316 and a CIDEr of 0.322, the best scores among the methods we compare against. On IU X-Ray, it attains a BLEU-4 of 0.201 and a CIDEr of 0.724, leading every baseline on every standard NLG metric. We further evaluate CogRad with RadGraph F1, CheXbert F1, and a hallucination analysis to assess clinical accuracy beyond standard text-overlap metrics, complemented by ablation studies and Grad-CAM-based visualizations that characterize each agent's contribution and the model's visual grounding.}
 

\maketitle

\section{Introduction}
\label{sec:intro}

Accurate and timely radiology reporting is essential for effective clinical decision-making and patient management. However, the growing volume of medical imaging studies has substantially increased the workload of radiologists, creating a demand for intelligent computer aided solutions. One important application of clinical artificial intelligence is the automatic generation of radiology reports from chest X-ray images. Hundreds of millions of chest radiographs are acquired worldwide each year, and the workload associated with manual interpretation contributes to inter-reader variability, diagnostic delays, and radiologist burnout. Automated RRG systems can serve as effective clinical decision-support tools by highlighting critical findings, reducing repetitive documentation tasks, and improving reporting efficiency and productivity.

Recent advances in vision language alignment and large language models (LLMs) have led to a new generation of RRG systems. R2GenCSR~\cite{ref3} leverages Mamba~\cite{ref6} for efficient linear complexity visual encoding with context-guided retrieval, R2GenGPT~\cite{ref2} couples a Swin Transformer~\cite{ref5} with LLaMA-2-7B for high-quality report generation, and AM-MRG~\cite{ref4} integrates dual Hopfield memory networks with a Swin Transformer to enable disease-aware associative reasoning. Despite their linguistic fluency, these approaches remain constrained by a single-pass generation paradigm: once visual features are encoded and projected into the LLM embedding space, reports are generated autoregressively with no mechanism for visual verification, uncertainty-driven attention redistribution, or iterative re-analysis of the image. As a result, generated reports may contain unsupported findings or overlook subtle abnormalities that would have benefited from additional visual scrutiny.

This workflow differs markedly from how radiologists work in practice. Rather than producing a report in a single pass, a radiologist first surveys the entire image to identify potential abnormalities, inspects suspicious regions more closely, progressively formulates findings, and checks each conclusion against the image before finalizing the report. Every stage of this process is visually grounded and evidence-driven, and it is not strictly sequential: radiologists frequently return to a region after an initial read, revising an impression once a later observation puts an earlier one in a different light. Information is continuously integrated across anatomical structures and imaging patterns, so that local findings are interpreted in light of the global clinical image rather than in isolation. It is this combination staged visual attention, explicit grounding of each claim in image evidence, and a willingness to revisit and revise that current single-pass RRG architectures do not represent explicitly.

A growing body of work~\cite{ref7,ref8,ref11} has explored structured multi-agent frameworks for radiology report generation as a way to make this staged reasoning explicit, decomposing report generation into multiple collaborating components to improve interpretability and enable intermediate verification. Existing approaches in this direction, however, each leave part of the radiologist's workflow unaddressed: some rely on text-only regional descriptions for inter-agent communication, discarding the continuous visual representation that would let a later stage re-examine the image directly~\cite{ref8}; some incur substantial inference cost from sequential tool invocations~\cite{ref7}; and some confine visual understanding to an isolated agent without a structured channel for sharing visual evidence across agents~\cite{ref11}. None of these directions provide a mechanism for the model to revisit its own output once a draft report has already been produced, which is precisely the self-checking step that distinguishes an experienced radiologist's reading from a single forward pass.

We introduce \textbf{CogRad}, a cognitively-inspired multi-agent framework for chest X-ray report generation that is organized around four stages of the radiologist's reading process global triage, focused investigation, structured reporting, and visual verification and that, unlike prior agentic approaches, carries a continuous visual representation through all four stages and is able to revisit a generated report when its own verification stage judges a statement to be weakly grounded. The four agents are as follows.

\subsection{Multi-Agent Cognitive Architecture}

The proposed framework, termed \textbf{CogRad}, employs a four-agent cognitive architecture designed to emulate the workflow of an experienced radiologist. Each agent performs a specialized task while collaboratively contributing to accurate disease classification and clinically reliable report generation.

\subsubsection*{Scout Agent: Dynamic Anatomical Scanner}

The Scout Agent performs an initial global assessment of the chest X-ray to predict the presence of 14 standard thoracic diseases and identify potentially abnormal regions. Unlike conventional region-partitioning approaches that rely on fixed image grids, the Scout Agent incorporates a \textit{Slot-Attention} mechanism. This module consists of multiple learnable slots that dynamically compete to represent clinically meaningful anatomical structures and pathological regions. Consequently, the model can automatically adapt to variations in patient anatomy, image size, and acquisition conditions.

The primary objective of the Scout Agent is to provide a coarse yet comprehensive understanding of the image by localizing abnormalities and generating a patient-specific anatomical representation that guides subsequent analysis.

\subsubsection*{Investigator Agent: Fine-Grained Diagnostic Analyzer}

The Investigator Agent receives the candidate abnormal regions identified by the Scout Agent and performs detailed examination of these areas. A self-attention mechanism enables interactions among different anatomical regions, allowing the model to capture clinically relevant dependencies. For example, automatically finding the boundaries of the lungs or heart dynamically, even if the image size changes.

To improve diagnostic precision, the Investigator Agent allocates additional computational resources to the top-$K$ most suspicious regions, enabling deeper feature extraction and more discriminative pathological representations. This targeted analysis provides detailed visual evidence for downstream report generation.

\subsubsection*{Writer Agent: Structured Clinical Reasoning Module}

The Writer Agent transforms the visual representations extracted by previous agents into a structured sequence suitable for report generation using a LLM, such as LLaMA-7B. Inspired by the reasoning process of radiologists, the information is organized hierarchically:

\begin{enumerate}
    \item Disease-level diagnostic context,
    \item Regional pathological evidence,
    \item Global image observations.
\end{enumerate}

Furthermore, a confidence-based gating mechanism suppresses tokens associated with highly unlikely disease categories. By filtering weak diagnostic signals before they reach the language model, the Writer Agent reduces the risk of hallucinated findings and improves report consistency.

\subsubsection*{Verifier Agent: Hallucination Detection and Correction}

The Verifier Agent serves as a fact-checking component that continuously validates the generated clinical report against image-derived evidence. Unlike existing approaches that utilize verification only during training, the proposed framework maintains the Verifier Agent during inference.

For each generated sentence, the Verifier Agent computes a confidence score within the range $[0,1]$. If the confidence falls below a predefined threshold (e.g., 0.5), the sentence is considered potentially unreliable. The system then initiates a corrective feedback cycle in which:

\begin{enumerate}
    \item The Scout Agent updates abnormality prioritization scores.
    \item The Investigator Agent performs additional examination of suspicious regions.
    \item The Writer Agent regenerates the report using refined visual evidence.
\end{enumerate}

This iterative verification process can be repeated up to two times, ensuring that the final report remains strongly grounded in image evidence and minimizing the occurrence of hallucinated clinical statements.

Our contributions are organized around four aspects of this design.

\begin{itemize}
    \item \textbf{A cognitively structured agent pipeline with learned visual regions:} We replace the fixed anatomical grouping used in prior region-aware RRG models with a slot-attention mechanism that discovers region representations directly from patch features, and we use triage-guided focus weighting and disease-gated visual prompting to carry this structure through to the language model's input, preserving visual-semantic alignment across the full pipeline rather than only at the point of image encoding.
    \item \textbf{An inference time self-correction loop:} Beyond using visual entailment as a training-time loss, we introduce a sentence-level verification mechanism that operates after a draft report has already been generated, allowing the model to flag and selectively regenerate statements it judges to be weakly grounded a capability closer to a radiologist's own self-checking than a single forward pass through the language model can offer.
    \item \textbf{A hallucination aware training signal without additional annotation:} The Verifier entailment loss gives the model a direct, differentiable signal for whether a claim is visually supported, reducing reliance on post-hoc filtering or extra clinical labels beyond the disease labels already used for Scout supervision.
    \item \textbf{Evaluation that goes beyond n-gram overlap:} Alongside the standard NLG metrics (BLEU-1/2/3/4, ROUGE-L, CIDEr) used, we also used RadGraph F1, CheXbert-14/5 F1, and an entity-level hallucination analysis, together with Grad-CAM-based interpretability and ablation studies, providing a more complete image of when the framework's gains do and do not translate into clinically accurate reporting.
\end{itemize}
\section{Related Work}
\label{sec:related}
RRG aims to automatically produce clinically accurate and coherent diagnostic reports from medical images, reducing radiologists' workload and improving reporting efficiency. Recent advances in deep learning, particularly vision-language models, have significantly improved report quality by jointly modeling visual findings and textual descriptions. More recently, multi-agent systems have emerged as a promising paradigm, enabling specialized agents to collaboratively perform image interpretation, clinical reasoning, and report refinement. This collaborative framework enhances report accuracy, consistency, and interpretability compared with conventional single-model approaches.
\subsection{Radiology Report Generation}
Early RRG methods adapted image captioning frameworks by integrating CNN-based feature extractors with attention mechanisms to generate free-text reports~\cite{ref1}. The introduction of Transformer based encoders, particularly the Swin Transformer~\cite{ref5}, substantially improved visual feature representation and report generation performance. Subsequent studies explored advanced retrieval and alignment strategies, including hierarchical vision-language alignment~\cite{ref9}, associative Hopfield memory networks~\cite{ref4}, and context-guided retrieval augmentation~\cite{ref3} to enhance disease-aware token retrieval. Among these approaches, R2Gen~\cite{ref1} established a strong baseline by introducing a memory-driven Transformer architecture for medical report generation.

A paradigm shift occurred with the emergence of LLMs. R2GenGPT~\cite{ref2} demonstrated that lightweight projection-based alignment of visual features with LLaMA-2-7B can match or even outperform specialized report-generation architectures. R2GenKG~\cite{ref10} further enhanced LLM-based report generation by incorporating knowledge graphs constructed with GPT-4o. Despite their strong performance, these approaches remain limited to a single forward pass, preventing iterative re-examination of visual evidence and verification of generated findings against the input image. To address this limitation, recent RRG frameworks have increasingly incorporated knowledge graphs to capture anatomical structures and their relational dependencies~\cite{ref10}. Despite their effectiveness, these approaches inadequately model hierarchical semantic dependencies that require the joint optimization of coarse-grained disease level consistency and fine-grained region-to-word alignment. Furthermore, they are computationally expensive and prone to error propagation from manually curated knowledge resources.

\subsection{Multi-Agent Systems for Medical AI}
The use of multi-agent frameworks for medical report generation remains relatively limited.~\cite{ref11} proposed a multi-agent system that decomposes RRG into specialized retrieval, drafting, refinement, visual analysis, and synthesis agents. However, the framework relies on sequential text-based coordination, and although a dedicated Vision Agent processes medical images, visual grounding is confined to a single stage without structured visual interaction across the remaining agents. In contrast, EviAgent~\cite{ref7} reformulates report generation as an evidence-driven, multi-step reasoning process by integrating a multimodal planner, discriminative perceptual tools, and pathology specific retrieval augmented generation (RAG). This design enables each diagnostic conclusion to be explicitly supported by corresponding visual evidence.
However, due to its multi-round tool invocation process, the framework relies on sequential single-agent planning, resulting in substantial inference latency. To improve laterality consistency and clinical accuracy, MARL-Rad~\cite{ref8} formulates RRG as a collaborative multi-agent reinforcement learning framework, where region-specific agents (left, right, and central) interact with a global integration agent and are jointly optimized using MA-GSPO with clinically verifiable reward signals. Nevertheless, inter-agent communication remains text-based, relying solely on intermediate regional diagnoses, without structured visual feature exchange or explicit cross-modal grounding. In contrast, the proposed CogRad enables communication through structured tensor representations, including triage scores, attention weights, and regional embeddings, while allowing each agent direct access to the original image features. This design facilitates richer information sharing and more effective cross-agent collaboration. 

\section{Methodology}
\label{sec:method}
\begin{figure}[h!]
\includegraphics[width=0.99\textwidth]{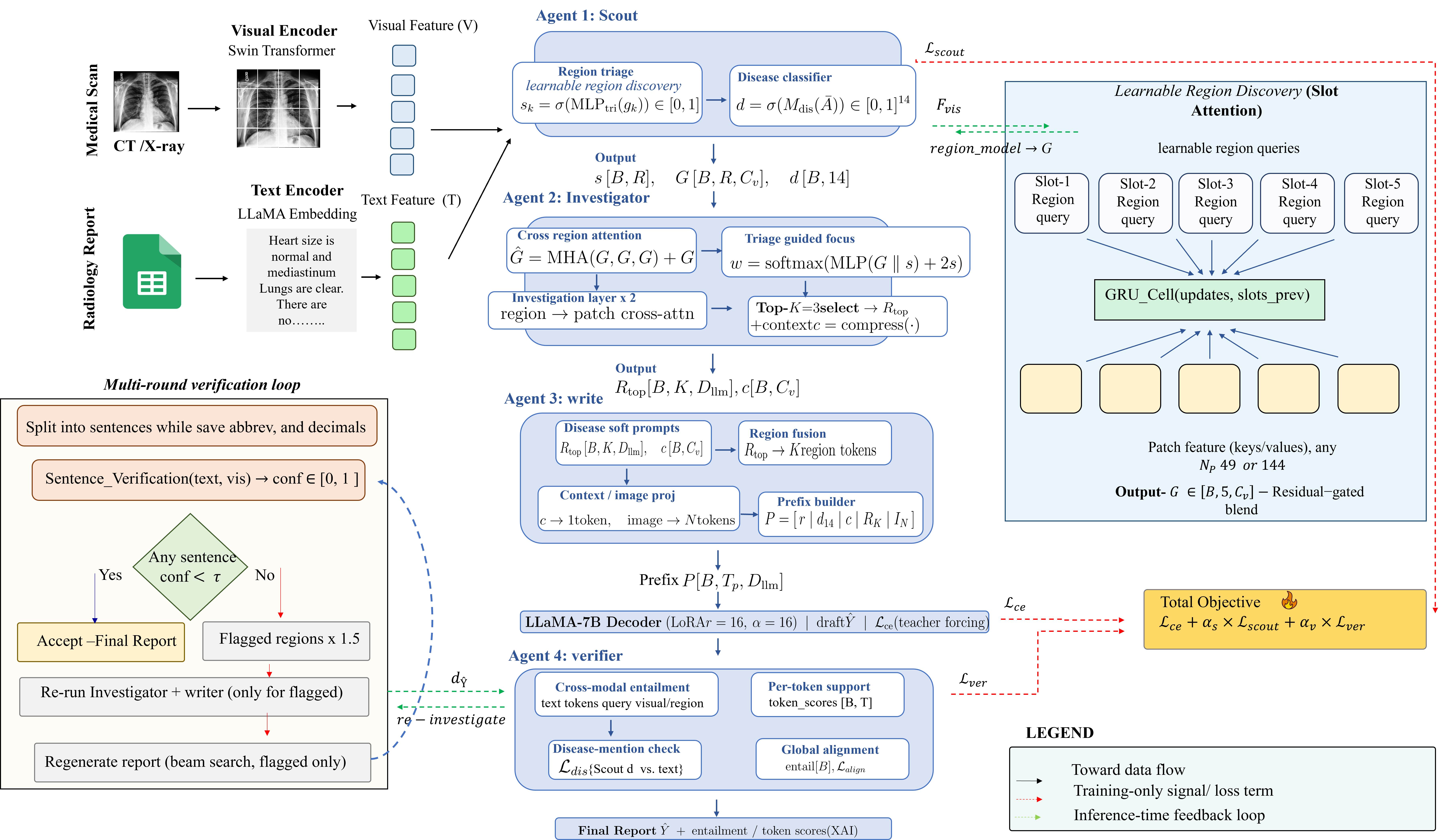}
\caption{Overview of the CogRad framework and its four specialized agents.}
\label{fig:architecture}
\end{figure}

We introduce CogRad, a cognitively inspired multi-agent framework for radiology report generation in which four specialized agents communicate through structured tensor representations rather than free-text instructions. The design is motivated by how an expert radiologist reads a chest X-ray: a rapid global survey to flag suspicious regions, a focused second look at those regions, a synthesis of findings into a structured impression, and a final self-check of whether each stated finding is actually supported by what is visible in the image. CogRad operationalizes this four-stage process as a \emph{Scout} agent, an \emph{Investigator} agent, a \emph{Writer} agent, and a \emph{Verifier} agent, all sharing a common vision encoder so that representations remain consistent across stages (Fig.~\ref{fig:architecture}). Unlike prior soft-prompting approaches to report generation, which condition the language model on a single pooled image embedding, CogRad constructs its visual prompt compositionally from disease-level, region-level, and patch-level evidence, and revisits that prompt at inference time whenever the language model own output casts doubt on its visual grounding.

\subsection{Problem Formulation}
 
Given a chest X-ray image $\mathbf{L} \in \mathbb{R}^{H \times W \times 3}$, where $H$ and $W$ denote the image height and width, the objective is to generate a clinically accurate diagnostic report $\mathbf{s} = (s_1, s_2, \ldots, s_T)$, where each $s_t$ is a discrete output token and $T$ is the report length. The training dataset is
\[
\mathcal{D} = \{(\mathbf{L}_x, \mathbf{s}_x)\}_{x=1}^{y},
\]
where $\mathbf{L}_x$ is the $x$-th chest X-ray image, $\mathbf{s}_x$ its corresponding ground-truth report, and $y$ the total number of image report pairs. To encourage anatomically grounded and factually consistent reports, we augment the standard language-modeling objective with two auxiliary clinical supervision losses, learning model parameters $\theta$ by
 
\begin{equation}
    \max_{\theta}\; \mathbb{E}_{(\mathbf{L}, \mathbf{s}) \sim \mathcal{D}} \Bigl[ \log p_\theta(\mathbf{s} \mid \mathbf{L}) \Bigr]
    \;-\; \lambda_s \mathcal{L}_{\text{Scout}}
    \;-\; \lambda_v \mathcal{L}_{\text{Ver}},
\end{equation}
where $\lambda_s, \lambda_v \in \mathbb{R}^{+}$ balance the disease-classification loss $\mathcal{L}_{\text{Scout}}$ of the Scout agent against the visual-entailment loss $\mathcal{L}_{\text{Ver}}$ of the Verifier agent, and $p_\theta(\mathbf{s} \mid \mathbf{L})$ is the model's conditional likelihood of the report given the image. The remainder of this section details each agent in turn.
 
\subsubsection{Vision Encoder}
A Swin Transformer-Base~\cite{ref5} serves as the shared visual backbone. For an input image $\mathbf{L}$, the encoder produces a grid of $N_p$ patch-level features,
\begin{equation}
    \mathbf{F}_{\text{vis}} = \text{Swin}(\mathbf{L}) \in \mathbb{R}^{B \times N_p \times C_v},
\end{equation}
where $C_v = 1024$ is the visual feature dimension and $B$ the batch size. We deliberately keep $N_p$ a free variable rather than a fixed constant: a $224\times224$ input yields $N_p=49$ patches, while higher-resolution inputs used for the CheXpert experiments yield $N_p=144$. All downstream modules, including the region-discovery mechanism described next, are designed to operate on an arbitrary number of patches without retraining or architectural modification. Patch features are additionally projected to the language model embedding dimension $C_l = 4096$ via a linear layer followed by layer normalization, giving $\mathbf{F}_{\text{llm}} \in \mathbb{R}^{B \times N_p \times C_l}$ for later use by the Writer.
 
\subsubsection{Scout Agent: Global Triage over Learned Anatomical Regions}
 
Rather than imposing a fixed anatomical grid on the patch grid which presupposes a particular input resolution and a hand-specified correspondence between patch index and anatomical location the Scout agent discovers its own set of region representations directly from the data. We adopt a slot-attention mechanism in which $|\mathcal{R}|=5$ learnable region queries compete, via iterative cross-attention, for the patch features that best explain them. Concretely, starting from a set of slot embeddings $\mathbf{U}_{\text{reg}} \in \mathbb{R}^{|\mathcal{R}| \times C_v}$ sampled from a learned Gaussian, each of $n_{\text{iter}}=3$ refinement steps computes attention between slot queries and patch keys/values, updates each slot with a GRU cell, and applies a per-slot feed-forward residual:
 
\begin{equation}
    \alpha_{kj} = \frac{\exp\!\left(\langle \mathbf{q}_k, \mathbf{k}_j \rangle / \sqrt{d}\right)}{\sum_{k'} \exp\!\left(\langle \mathbf{q}_{k'}, \mathbf{k}_j \rangle / \sqrt{d}\right)}, \qquad
    \mathbf{u}_k = \sum_{j=1}^{N_p} \frac{\alpha_{kj}}{\sum_{j'} \alpha_{kj'}}\, \mathbf{v}_j,
\end{equation}
where $\mathbf{q}_k$, $\mathbf{k}_j$, and $\mathbf{v}_j$ are linear projections of slot $k$ and patch $j$, $d$ is the dimension of these projections used to scale the dot product, and the competition for normalization is over slots rather than patches, ensuring that each region of the image is claimed by exactly one slot in expectation. Because patch position carries diagnostic information (e.g., the lower zones are more likely to show effusions than the apices), we add a positional bias to the patch features before slot attention. This bias is stored as a base embedding at a reference resolution and bilinearly interpolated to match $N_p$ at runtime, allowing the same learned positional prior to transfer across the differing input resolutions of our two datasets. The resulting slot outputs are projected back to $C_v$ and lightly blended with the global mean-pooled patch feature through a learnable per-slot gate, yielding region features $\mathbf{H} \in \mathbb{R}^{B \times |\mathcal{R}| \times C_v}$ with the same shape as the fixed grouping it replaces, so that downstream agents require no modification. We also retain the slot-to-patch attention maps produced in the final iteration, which double as interpretable region-assignment maps at inference time.
 
Each region representation $\mathbf{h}_k$ is then scored for abnormality by a sigmoid-gated MLP,
\begin{equation}
    r_k = \sigma\!\left(\text{MLP}_{\text{triage}}(\mathbf{h}_k)\right) \in [0,1],
\end{equation}
producing the per-region triage vector $\mathbf{r} = (r_1, \ldots, r_{|\mathcal{R}|})$ that drives the Investigator agent. In parallel, a 14-class disease classifier operates on the global average patch feature $\bar{\mathbf{a}} = \text{mean}(\mathbf{F}_{\text{vis}}) \in \mathbb{R}^{B \times C_v}$ to predict soft pathology probabilities $\mathbf{c} \in [0,1]^{B \times 14}$, supervised by labels $\mathbf{y} \in \{0,1\}^{B \times 14}$ extracted from the reference reports via negation-aware keyword matching:
\begin{equation}
    \mathbf{c} = \sigma\!\left(\text{MLP}_{\text{dis}}(\bar{\mathbf{a}})\right), \qquad
    \mathcal{L}_{\text{Scout}} = \text{BCE}(\mathbf{c}, \mathbf{y}).
\end{equation}
The Scout agent passes the triage scores $\mathbf{r}$, the disease probabilities $\mathbf{c}$, and the region features $\mathbf{H}$ forward to the Investigator.
 
\subsubsection{Investigator Agent: Adaptive Regional Focus}
 
The Investigator concentrates representational capacity on the regions the Scout flags as suspicious, in the same way a radiologist returns for a closer look at an area of concern rather than re-examining the whole image uniformly. The region features first pass through a self-attention layer in which $\mathbf{H}$ serves simultaneously as queries, keys, and values, allowing regions to inform one another and capture clinically meaningful co-occurrence patterns for instance, the frequent association between cardiomegaly and pleural effusion:
\begin{equation}
    \hat{\mathbf{H}} = \text{MHA}(\mathbf{H}, \mathbf{H}, \mathbf{H}) + \mathbf{H}.
\end{equation}
Triage-guided focus weights are then computed by concatenating each region interacted representation with its scalar triage score and passing the result through an MLP, with the triage score additionally injected as a direct additive bias so that high-scoring regions are favored even before the MLP has fully adapted:
\begin{equation}
    \boldsymbol{\phi} = \text{softmax}\!\left(\text{MLP}\bigl([\hat{\mathbf{H}}; \mathbf{r}]\bigr) + 2\mathbf{r}\right) \in \mathbb{R}^{B \times |\mathcal{R}|}.
\end{equation}
Across $L=2$ investigation layers, each region representation is iteratively refined by cross-attending to the full patch grid, with the resulting update scaled by its focus weight before being added back:
\begin{equation}
    \mathbf{H}^{(\ell+1)} = \text{LN}\!\left(\mathbf{H}^{(\ell)} + \boldsymbol{\phi} \odot \text{MHA}\bigl(\mathbf{H}^{(\ell)}, \mathbf{F}_{\text{vis}}, \mathbf{F}_{\text{vis}}\bigr)\right), \qquad \ell = 0, \ldots, L-1,
\end{equation}
where $\odot$ denotes elementwise multiplication along the region axis. This scheme allocates more refinement to regions Scout considers abnormal without discarding information from regions it considers normal, since all regions continue to receive at least a baseline residual update. After the final layer, the top-$K$ regions by focus weight ($K=3$) are selected and projected into the language model embedding space, $\mathbf{R}_{\text{top}} \in \mathbb{R}^{B \times K \times C_l}$, while a focus-weighted summary of all regions is compressed into a single global investigation context vector $\mathbf{c}_{\text{inv}} \in \mathbb{R}^{B \times C_v}$. Both are passed to the Writer.
 
\subsubsection{Writer Agent: Disease-Gated Visual Prefix}
 
Before any text is generated, the Writer assembles the Scout's and Investigator outputs into a single structured visual prefix that conditions the language model. The token ordering is a deliberate design choice intended to mirror a radiologist's reasoning: disease-level prior context is presented first, followed by the specific regional findings that support or contradict it, and finally the full image as a basis for confirmation.
 
Disease scores are converted into soft prompt tokens through a sigmoid gate conditioned on the Scout score for that disease, so that diseases predicted as absent are driven toward a near-zero embedding and are effectively masked from the language model without being removed from the sequence:
\begin{equation}
    \mathbf{p}_k = \mathbf{e}_k \odot \sigma\!\left(\mathbf{V}_{\text{gate}}\, c_k + \mathbf{b}\right), \qquad k = 1, \ldots, 14,
\end{equation}
where $\mathbf{e}_k \in \mathbb{R}^{C_l}$ is a learnable embedding for disease $k$, $\mathbf{V}_{\text{gate}} \in \mathbb{R}^{C_l \times 1}$ projects the scalar disease score into a full-dimensional gate, and $\mathbf{b}$ is a learnable bias. The full visual prefix concatenates a learnable prefix token, the 14 disease tokens, the projected investigation context, the top-$K$ regional tokens, and the projected global patch features:
\begin{equation}
    \mathbf{Q} = \bigl[\mathbf{q}_{\text{cls}};\, \mathbf{p}_{1:14};\, \mathbf{c}_{\text{inv}}';\, \mathbf{R}_{\text{top}};\, \mathbf{F}_{\text{llm}}\bigr] \in \mathbb{R}^{B \times T_P \times C_l}, \qquad T_P = 1 + 14 + 1 + K + N_p,
\end{equation}
where $\mathbf{c}_{\text{inv}}'$ is the investigation context projected to $C_l$. Note that $T_P$ depends on $N_p$ and therefore on input resolution: it is $68$ tokens for the $224\times224$ inputs used on IU X-Ray and $163$ tokens for the $384\times384$ inputs used on CheXpert Plus ($N_p = 144$), with no change to the architecture itself. The language model's input sequence during training is $\mathbf{X} = [\mathbf{x}_{\text{bos}}; \mathbf{Q}; \mathbf{E}_{\text{text}}]$, where $\mathbf{E}_{\text{text}}$ are the embeddings of the target report tokens, and the model is trained with the standard autoregressive cross-entropy loss
\begin{equation}
    \mathcal{L}_{\text{CE}} = -\sum_{t=1}^{T} \log\, p_\theta(s_t \mid \mathbf{X}_{<t}).
\end{equation}
 
\subsubsection{Verifier Agent: Training Time Visual Entailment and Inference Time Re-Examination}
 
Report generators conditioned only on a forward language-modeling loss have no explicit mechanism to penalize a claim that is fluent but visually unsupported. We address this with a Verifier agent that operates in two complementary ways: as a training-time auxiliary loss that discourages ungrounded generation, and as an inference-time controller that detects and revises low-confidence claims in the model's own draft output.
 
\paragraph{Training time entailment loss.} During training, the patch features $\mathbf{F}_{\text{vis}}$ and Scout's region features $\mathbf{H}$ are projected into a shared verification space of dimension $C_h = 512$, alongside the target text embeddings:
\begin{equation}
    \mathbf{Z} = \text{MLP}_z\!\left([\mathbf{F}_{\text{vis}}; \mathbf{H}]\right) \in \mathbb{R}^{B \times (N_p + |\mathcal{R}|) \times C_h}, \qquad
    \mathbf{G} = \text{MLP}_g(\mathbf{E}_{\text{text}}) \in \mathbb{R}^{B \times T \times C_h}.
\end{equation}
Each text token attends to the visual keys via cross-attention, $(\mathbf{O}, \boldsymbol{\beta}) = \text{MHA}(\mathbf{G}, \mathbf{Z}, \mathbf{Z})$, producing both an attended visual context $\mathbf{O}$ and attention weights $\boldsymbol{\beta} \in \mathbb{R}^{B \times T \times (N_p+|\mathcal{R}|)}$ that can be read directly as token-to-patch grounding maps at no extra inference cost. A per-token entailment logit $\hat{e}_t = \text{MLP}_e([\mathbf{g}_t; \mathbf{o}_t])$ is trained toward a high target, encouraging every generated token to be visually defensible, while a disease-mention consistency term checks whether diseases the Scout scored highly are actually reflected in the attended representation, and a global alignment term encourages the mean-pooled visual and textual representations to agree:
\begin{equation}
    \mathcal{L}_{\text{Ver}} =
    \underbrace{\text{BCE}(\hat{\mathbf{e}}, \mathbf{1}\!\cdot\!0.9)}_{\text{visual grounding}}
    \;+\; 0.3\, \underbrace{\text{MSE}\!\left(\sigma(\hat{\mathbf{c}}_m), \mathbf{c}_{\text{sg}}\right)}_{\text{disease consistency}}
    \;+\; 0.2\, \underbrace{\bigl(1 - \cos(\bar{\mathbf{z}}, \bar{\mathbf{g}})\bigr)}_{\text{global alignment}},
\end{equation}
where the pooled representations and the disease-mention prediction are
\begin{equation}
    \bar{\mathbf{z}} = \frac{1}{N_p + |\mathcal{R}|}\sum_{i} \mathbf{z}_i, \qquad
    \bar{\mathbf{g}} = \frac{1}{T}\sum_{t=1}^{T} \mathbf{g}_t, \qquad
    \hat{\mathbf{c}}_m = \text{MLP}_m\!\Bigl(\tfrac{1}{T}\textstyle\sum_{t=1}^{T} \mathbf{o}_t\Bigr) \in \mathbb{R}^{B \times 14},
\end{equation}
and $\mathbf{c}_{\text{sg}}$ denotes the Scout disease scores with gradients stopped, so that the Verifier's loss shapes the language model's outputs without back-propagating into the Scout's own classification head. The Verifier adds no parameters to the inference-time forward pass through the language model and is active only during training in this role.

\paragraph{Inference time multi-round re-examination:} A single forward pass through Writer and the language model produces a fluent draft, but fluency is not evidence of grounding, particularly for findings that are plausible in general chest radiographs but absent from the specific image at hand. We therefore extend the Verifier with a lightweight sentence-level head that, at inference time, re-examines the model's own draft sentence by sentence and selectively triggers regeneration where confidence is low.

Concretely, the draft report is split into sentences using a rule-based splitter that protects common clinical abbreviations and decimal values from being mistaken for sentence boundaries. For each sentence, its tokens are embedded and mean-pooled, projected into the verification space, and concatenated with the mean-pooled visual context (patch and region features, pooled identically to training time) before being passed through a sentence-confidence head:
\begin{equation}
    \text{conf}(\text{sentence}) = \sigma\!\left(\text{MLP}_{\text{sent}}\bigl([\bar{\mathbf{u}}_{\text{sent}}; \bar{\mathbf{k}}]\bigr)\right) \in [0,1].
\end{equation}
Sentences scoring below a confidence threshold $\tau$ (we use $\tau = 0.5$) flag their report for re-examination. For flagged samples, the Scout's triage scores are uniformly boosted ($\mathbf{s} \leftarrow \min(1.5\,\mathbf{s}, 1)$) before being passed back through the Investigator and Writer, producing an updated visual prefix that allocates more attention to potentially under-examined regions; the language model then regenerates the full report for that sample from the updated prefix. This loop runs for up to two rounds and exits early once no sentence in any sample remains flagged, so that well-grounded drafts incur no additional inference cost beyond a single forward pass of the (computationally cheap) sentence scorer. We deliberately regenerate the full report for a flagged sample rather than splicing in a replacement for only the flagged sentence, since chest X-ray findings are frequently described in a way that depends on surrounding sentence context (e.g., a stability statement that refers back to an earlier finding), and partial replacement risks introducing internal inconsistency that a full regeneration from a corrected prefix avoids.

\subsection{Training Objective}

CogRad is trained end-to-end by minimizing
\begin{equation}
     \mathcal{L} = \mathcal{L}_{\text{CE}} + \lambda_s\, \mathcal{L}_{\text{Scout}} + \lambda_v\, \mathcal{L}_{\text{Ver}},
    \label{eq:total_loss}
\end{equation}
where $\mathcal{L}_{\text{CE}}$ is the autoregressive report-generation loss, $\mathcal{L}_{\text{Scout}}$ the disease-classification loss, and $\mathcal{L}_{\text{Ver}}$ the visual-entailment loss, with $\lambda_s$ and $\lambda_v$ scalar weights kept smaller than the coefficient on $\mathcal{L}_{\text{CE}}$ so that clinical supervision shapes generation without compromising fluency. The sentence-level confidence head used by the inference-time re-examination loop shares the Verifier's visual and textual projections and is trained jointly with the rest of the Verifier; it adds negligible parameter overhead and does not introduce a separate loss term beyond the entailment objective above. All four agents and the language model backbone (LLaMA-7B, optionally adapted with LoRA on the vision encoder, the language model, or both) are optimized jointly, with the region-discovery mechanism in the Scout agent receiving gradient signal both from the disease-classification loss and, indirectly, from the downstream report-generation loss.

\subsection{TRIPOD Compliance}
\label{subsec:tripod}

This study adheres to the TRIPOD (Transparent Reporting of a multivariable prediction model for Individual Prognosis Or Diagnosis) reporting guidelines for prediction model development and validation. Table~\ref{tab:tripod} summarizes compliance with the TRIPOD items relevant to this work.

\begin{table}[!htbp]
\caption{TRIPOD compliance checklist for CogRad.
Items not applicable to radiology report generation
are marked N/A.}
\label{tab:tripod}
\begin{tabular*}{\textwidth}{@{\extracolsep\fill}clp{5.2cm}c}
\toprule
\textbf{Item} & \textbf{Section} & \textbf{TRIPOD Criterion} & \textbf{Status} \\
\midrule
1  & Title        & Identify the study as developing a prediction model          & \checkmark \\
2  & Abstract     & Structured summary of objectives, methods, and results       & \checkmark \\
3  & Introduction & Scientific background and clinical motivation                & \checkmark \\
4  & Introduction & Objectives and intended use of the model                     & \checkmark \\
5  & Methods      & Source of data and eligibility criteria                      & \checkmark \\
6  & Methods      & Outcome definition (radiology report generation from CXR)    & \checkmark \\
7  & Methods      & Predictors and pre-processing described                      & \checkmark \\
8  & Methods      & Sample size and data partitioning reported                   & \checkmark \\
9  & Methods      & Statistical and model development methods described          & \checkmark \\
10 & Methods      & Model evaluation metrics defined (BLEU, ROUGE-L, CIDEr, RadGraph F1, CheXbert F1, hallucination rate) & \checkmark \\
11 & Methods      & Internal validation strategy (held-out test set)             & \checkmark \\
12 & Results      & Characteristics of study datasets reported                   & \checkmark \\
13 & Results      & Model performance reported on CheXpert Plus and IU X-Ray    & \checkmark \\
14 & Results      & Ablation study and comparison with SOTA provided             & \checkmark \\
15 & Results      & Explainability analysis via GradCAM heatmaps provided        & \checkmark \\
16 & Discussion   & Limitations discussed                                        & \checkmark \\
17 & Discussion   & Generalisability and future directions addressed             & \checkmark \\
18 & Other        & Funding and conflicts of interest declared                   & \checkmark \\
\bottomrule
\end{tabular*}
\end{table}

\section{Experiments and Analyses}
\label{sec:experiments}

\subsection{Datasets}

We evaluate CogRad on two publicly available chest X-ray report generation benchmarks that differ substantially in scale, imaging protocol, and report structure, allowing us to probe how the framework behaves both in a large, clinically diverse setting and in a smaller, more curated one.

The CheXpert Plus dataset~\cite{ref13} comprises 223,462 paired chest X-ray images and radiology reports collected from 187,711 examinations involving 64,725 patients. Each report is annotated with 14 expert-labeled disease categories following the CheXpert labeling schema. To prevent patient-level leakage across partitions, we use a patient-wise split of 70\% training, 10\% validation, and 20\% testing.

\begin{figure*}[h]
\centering
\begin{minipage}[]{10cm}
  \centering
  \includegraphics[width = 10cm]{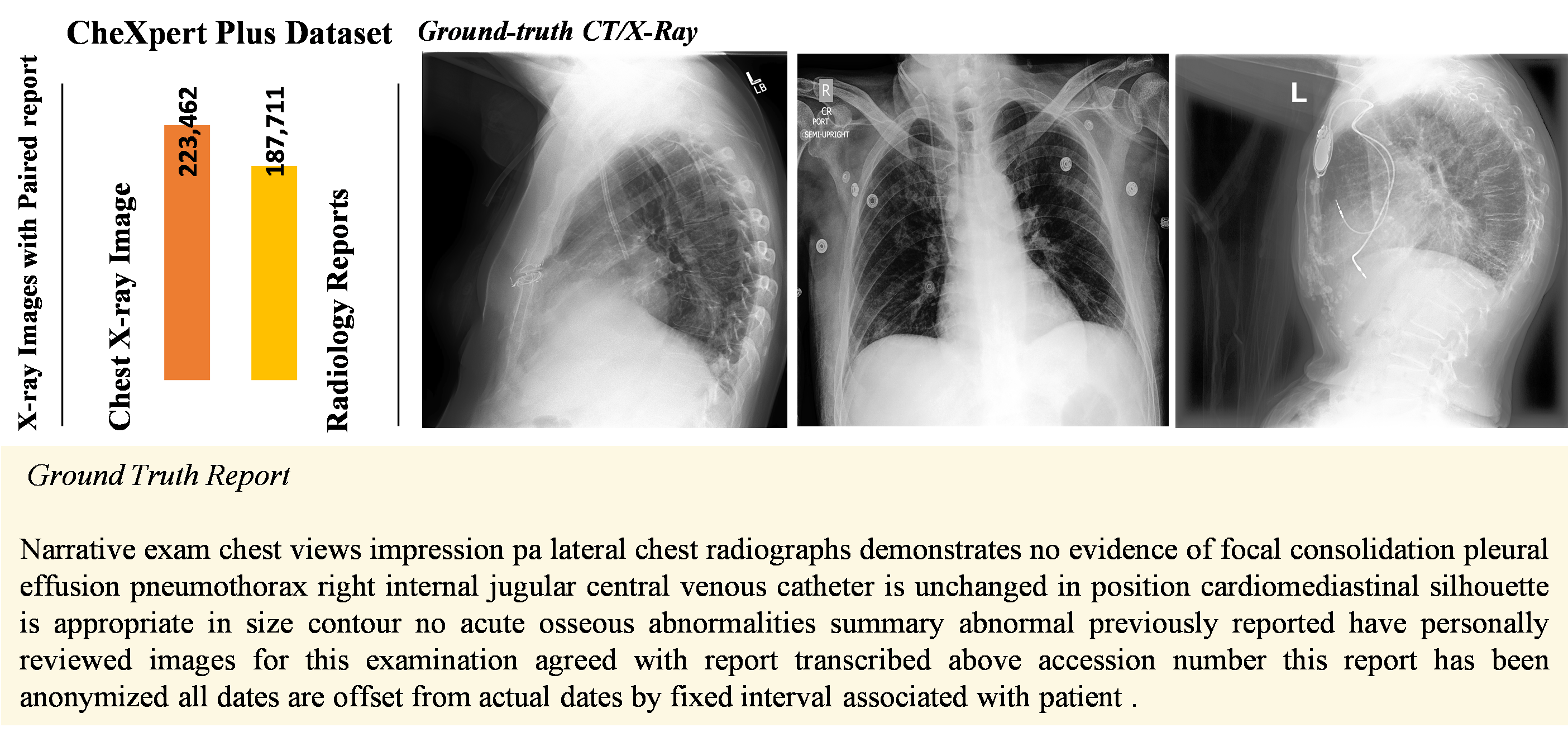}
  \vspace{-0.5cm}
    \begin{center}
    \textbf{CheXpert Plus}
    \end{center}
\end{minipage}
\begin{minipage}[]{10cm}
  \centering
  \includegraphics[width = 10cm]{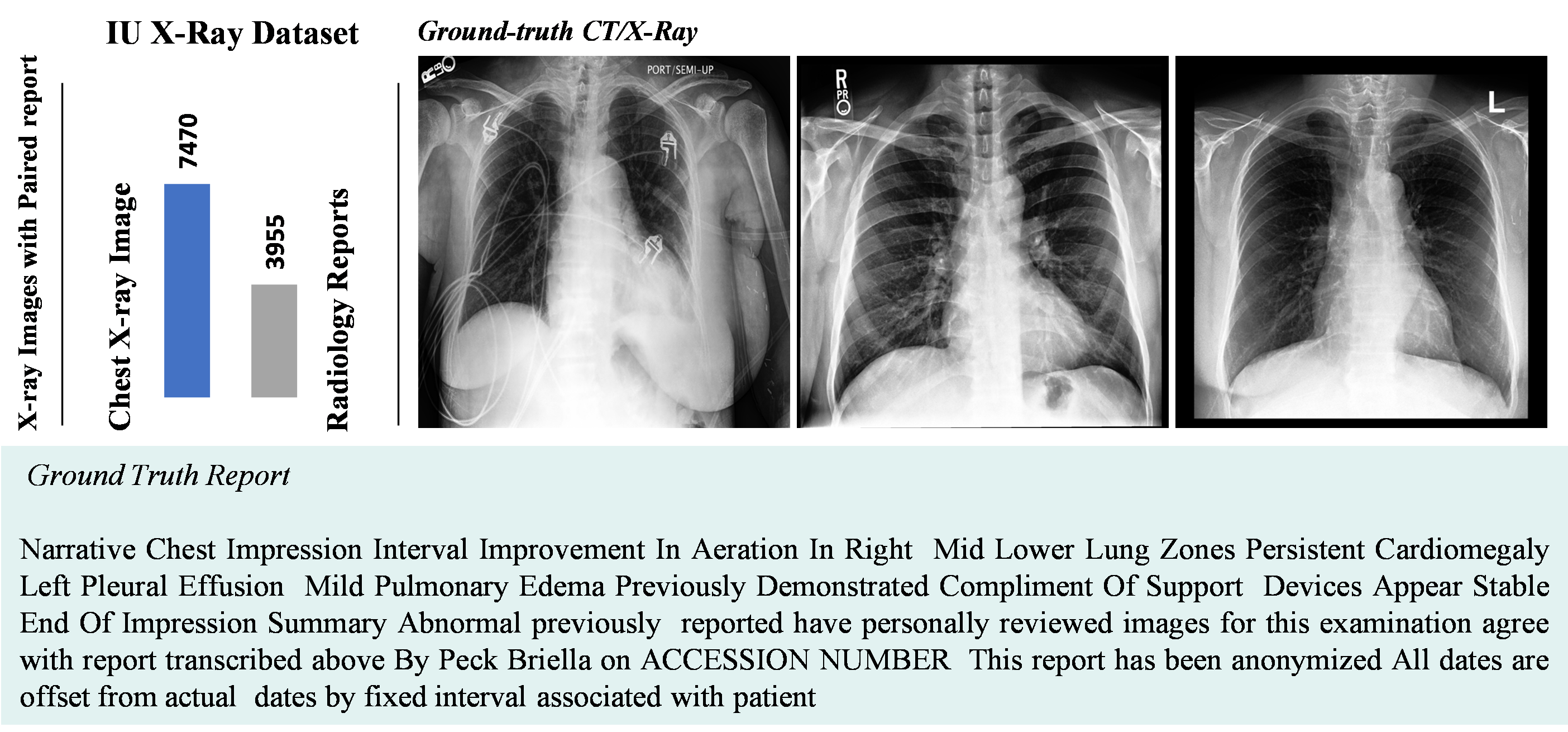}
  \vspace{-0.5cm}
    \begin{center}
    \textbf{IU X-Ray}
    \end{center}
\end{minipage}
  \caption{Overview of the two evaluation datasets.}
  \label{figDS}
\end{figure*}

IU X-Ray~\cite{ref14} is a smaller, publicly available dataset from Indiana University containing 3,955 radiology reports and 7,470 chest X-ray images, each report consisting of \emph{Findings} and \emph{Impression} sections. We follow the official train/validation/test split and evaluate on both sections according to the standard protocol. An overview of both datasets is given in Fig.~\ref{figDS}.

\subsection{Implementation Details}
CogRad uses LLaMA-2-7B-Chat~\cite{ref15} as the backbone language model, fine-tuned with LoRA~\cite{ref16} ($r=16$, $\alpha=16$), and a Swin Transformer-Base~\cite{ref5} vision encoder, producing $N_p=49$ patch tokens at the $224\times224$ resolution used for IU X-Ray and $N_p=144$ at the higher resolution used for CheXpert Plus, with no change to the architecture between the two settings. All agent modules use a hidden dimension of $512$, $|\mathcal{R}|=5$ learned region slots, $K=3$ selected regions, and $L=2$ investigation layers. Training was run on a single NVIDIA A100 GPU with AdamW, cosine annealing, and bf16 mixed precision; CheXpert Plus was trained for up to 25 epochs with learning rate $5\times10^{-6}$, 8-step gradient accumulation, and clinical loss weights $\lambda_s=0.1$, $\lambda_v=0.2$, while IU X-Ray was trained for up to 25 epochs with learning rate $1\times10^{-4}$, 4-step accumulation, and $\lambda_s=0.1$, $\lambda_v=0.1$; in both cases the checkpoint with the best combined BLEU-4 and CIDEr on the validation set was kept (epoch 18 for CheXpert Plus, epoch 9 for IU X-Ray). Inference uses beam search with beam size 5 for CheXpert Plus and beam size 3 for IU X-Ray, gradient clipping at 1.0, and up to two rounds of the inference-time re-examination loop with sentence-confidence threshold $\tau=0.5$.

\subsection{Evaluation Metrics}
We assess generated reports along two complementary axes. Linguistic quality is measured with the standard NLG metrics used throughout the report-generation literature: BLEU-1/2/3/4~\cite{ref17}, ROUGE-L~\cite{ref18}, and CIDEr~\cite{ref19}, computed at the corpus level. Clinical accuracy is assessed with three complementary metrics: RadGraph F1~\cite{jain2021radgraph}, which scores entity and relation level overlap between generated and reference reports using a clinically trained information extraction model; CheXbert F1~\cite{smit2020combining}, computed both over the full 14-label CheXpert taxonomy and over the 5 competition conditions, using a BERT based labeler applied independently to generated and reference text and an entity level hallucination analysis that extracts pathology and device mentions from both generated and reference reports via a negation-aware lexicon, reporting the hallucination rate (fraction of generated entities absent from the reference), the miss rate (fraction of reference entities absent from the generation), and their harmonic mean as an entity F1 score. For interpretability, we visualize the Verifier agent's cross-modal attention weights as GradCAM heatmaps and inspect the confidence scores produced by the sentence-level verification head used in the inference-time re-examination loop.

\subsection{Comparison with State-of-the-Art}
\label{sec:sota}

Table~\ref{tab:main_results} compares CogRad against recent report-generation baselines on both datasets. On CheXpert Plus, CogRad obtains substantially higher BLEU-1 through BLEU-4 and ROUGE-L than every baseline we compare against, with BLEU-4 reaching 0.316, well ahead of any baseline in the comparison. We attribute this gap primarily to the disease-gated visual prefix, which conditions generation on an explicit, per-disease probability rather than a single pooled image vector, encouraging the language model to commit early to a small set of likely findings and phrase them consistently a pattern that n-gram overlap metrics reward heavily. CogRad also obtains the best CIDEr score in the comparison, reaching 0.322, indicating that the disease-gated visual prefix gains are not confined to n-gram precision metrics alone. We examine whether this NLG advantage is matched by entity-level clinical accuracy in Section~\ref{sec:hallucination}.

On IU X-Ray, CogRad outperforms every baseline on every NLG metric, reaching a BLEU-4 of 0.201 and a CIDEr of 0.724. Given that IU X-Ray's reports are shorter and more templated than CheXpert Plus's, this result indicates that the gains from CogRad's explicit triage-investigate-write-verify decomposition are not confined to larger, more heterogeneous corpora. We examine whether this NLG advantage is matched by entity-level clinical accuracy in Section~\ref{sec:hallucination}.


\begin{table}[!htbp]
\caption{Comparison with state-of-the-art methods on CheXpert
Plus and IU X-Ray. Best results in \textbf{bold}, second best
\underline{underlined}. -- indicates not reported in the
original paper.}
\small
\label{tab:main_results}
\centering
\begin{tabular}{l l ccccc}
\toprule
\textbf{Dataset} & \textbf{Method}
& \textbf{B-1} & \textbf{B-2} & \textbf{B-4}
& \textbf{R-L} & \textbf{CIDEr}  \\
\midrule
\multirow{6}{*}{\textbf{CheXpert Plus}}
& R2GenCSR~\cite{ref3}~(2024) & --    & --    & 0.103 & 0.272 & \underline{0.193}   \\
& AM-MRG~\cite{ref4}~(2025)    & 0.381 & 0.238 & 0.109 & 0.282 & 0.221  \\
& R2GenKG~\cite{ref10}~(2025)   & 0.376 & 0.234 & 0.106 & 0.269 & 0.125 \\
& EMRRG~\cite{ref20}~(2025)     & 0.375 & 0.232 & 0.104 & 0.273 & 0.167  \\
\cmidrule{2-7}
&    
\textbf{CogRad (Ours)}       & \textbf{0.429} & \textbf{0.366} & \textbf{0.316} & \textbf{0.480} & \textbf{0.322}  \\
\midrule
\multirow{6}{*}{\textbf{IU X-Ray}}
& R2Gen~\cite{ref1}~(2020)     & 0.470 & 0.304 & 0.165 & 0.371 & --    \\
& R2GenGPT~\cite{ref2}~(2023)  & \underline{0.488} & 0.316 & 0.173 & 0.377 & 0.438  \\
& AM-MRG~\cite{ref4}~(2025)    & 0.489 & \underline{0.339} & 0.192 & \underline{0.384} & 0.613  \\
& HSA~\cite{ref9}~(2024)       & 0.527 & 0.361 & 0.196 & 0.405 & 0.598  \\
& R2GenKG~\cite{ref10}~(2025)   & 0.468 & 0.312 & \underline{0.181} & 0.383 & 0.701  \\
\cmidrule{2-7}
&\textbf{CogRad (Ours)}       & \textbf{0.549} & \textbf{0.379} & \textbf{0.201} & \textbf{0.429} & \textbf{0.724} \\
\bottomrule
\end{tabular}
\end{table}

\subsection{Ablation Study}
\label{sec:ablation}
To isolate the contribution of each agent auxiliary loss, we re-train CogRad with the Scout agent disease-classification supervision removed (\emph{w/o Scout}), with the Verifier visual-entailment loss removed (\emph{w/o Verifier}), and, on IU X-Ray, with both removed simultaneously (\emph{w/o Scout+Verifier}). 


\begin{table}[!htbp]
\caption{Ablation study on CheXpert Plus and IU X-Ray. All scores are corpus-level NLG metrics on the respective test sets.}
\small
\label{tab:ablation}
\centering
\begin{tabular}{l l cccc}
\toprule
\textbf{Dataset} & \textbf{Variant} & \textbf{B-1} & \textbf{B-4} & \textbf{R-L} & \textbf{CIDEr} \\
\midrule
\multirow{3}{*}{\textbf{CheXpert Plus}}
& Full CogRad                       & \textbf{0.429} & \textbf{0.316} & \textbf{0.480} & \textbf{0.322} \\
& w/o Scout                         & 0.420 & 0.312 & 0.478 & 0.289 \\
& w/o Verifier                      & 0.427 & 0.310 & 0.478 & 0.273 \\
\midrule
\multirow{4}{*}{\textbf{IU X-Ray}}
& Full CogRad                       & \textbf{0.549} & \textbf{0.201} & \textbf{0.429} & \textbf{0.724} \\
& w/o Scout      & 0.449 & 0.189 & 0.404 & 0.576 \\
& w/o Verifier                      & 0.458 & 0.175 & 0.421 & 0.605 \\
& w/o Scout+Verifier & 0.471 & 0.186 & 0.410 & 0.655 \\
\bottomrule
\end{tabular}
\end{table}

\subsection*{Analysis} On both datasets, the full model outperforms every ablated variant on every metric. On CheXpert Plus, removing the Verifier entailment loss costs slightly more on CIDEr than removing the Scout's disease-classification loss (0.273 vs.\ 0.289), while both ablations produce smaller, consistent reductions on BLEU-1, BLEU-4, and ROUGE-L. On IU X-Ray, the margins are larger and the pattern reverses: removing the Scout costs more on CIDEr than removing the Verifier (0.576 vs.\ 0.605), and removing both components together leaves CIDEr at 0.655, higher than either single ablation alone. This asymmetry between datasets the Verifier mattering more on CheXpert Plus and the Scout mattering more on IU X-Ray suggests that the two losses provide complementary rather than redundant signal, with their relative importance shaped by dataset-specific factors such as label density and report length rather than by a fixed hierarchy between the two components.

\begin{figure}[h!]
\centering
\includegraphics[width=0.85\textwidth]{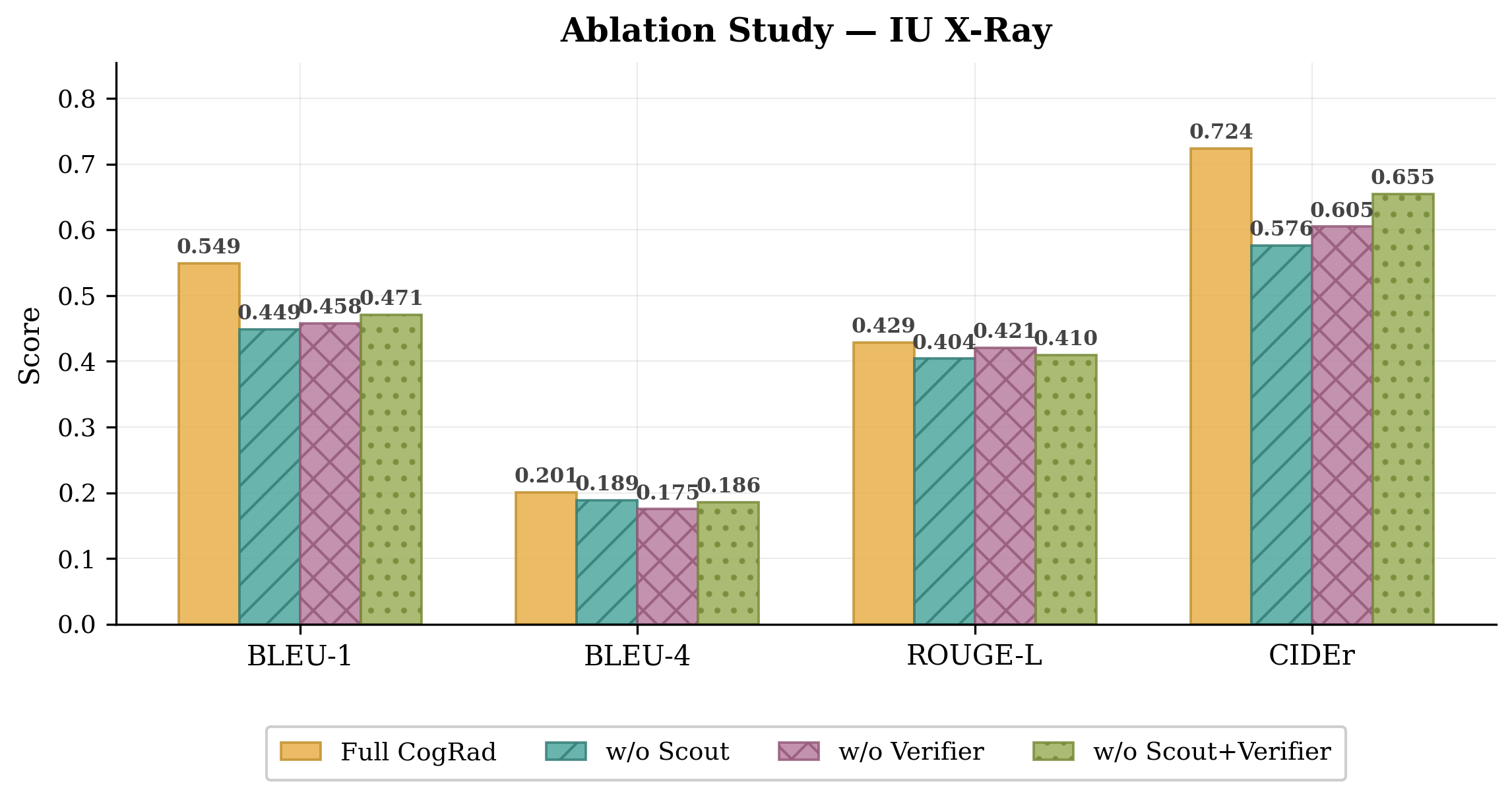}
\caption{Ablation results on IU X-Ray.}
\label{fig:ablation_bars_iu}
\end{figure}

\begin{figure}[h!]
\centering
\includegraphics[width=0.85\textwidth]{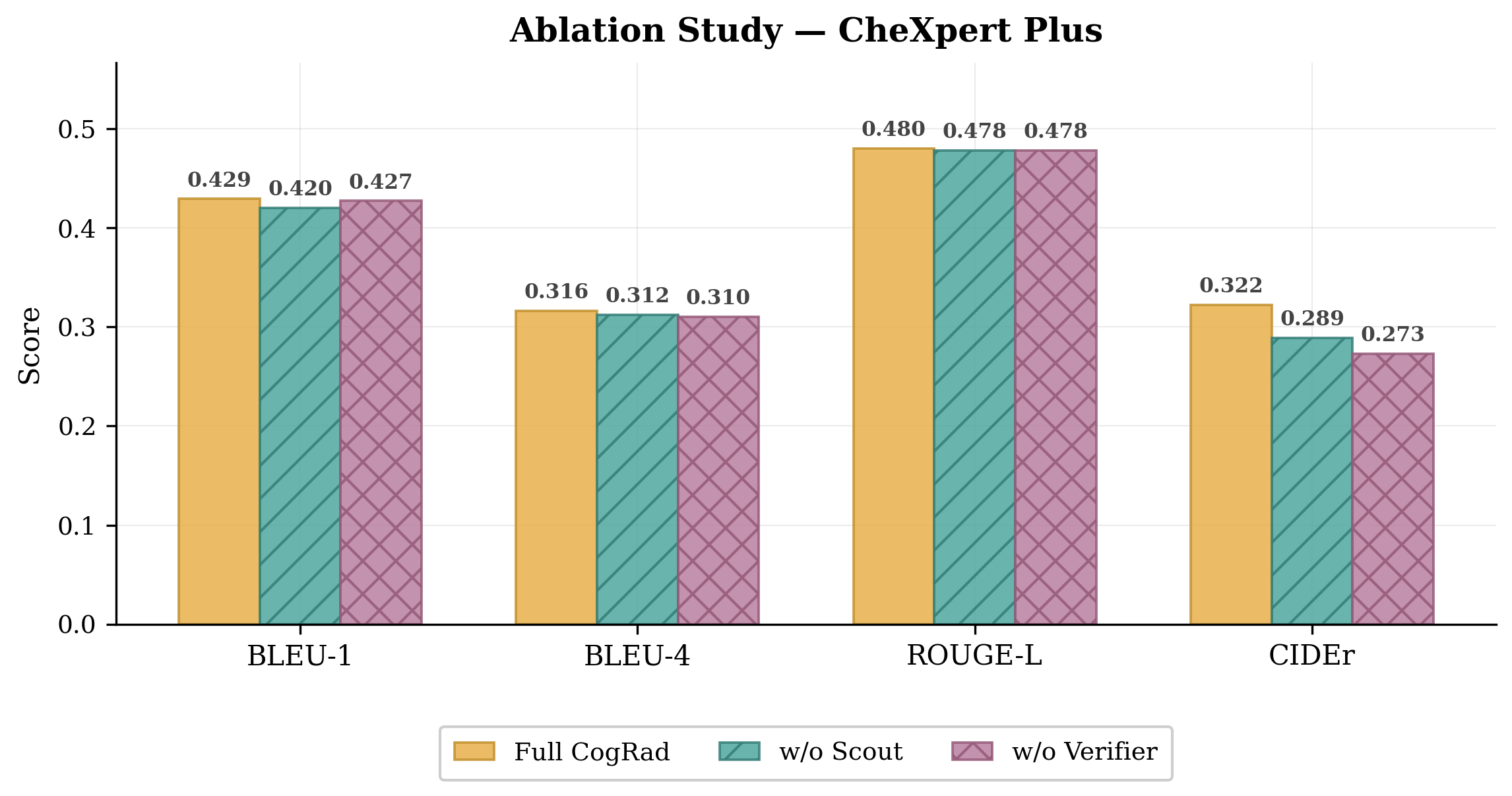}
\caption{Ablation results on CheXpert Plus. All three runs are independently trained.}
\label{fig:ablation_bars}
\end{figure}

\begin{figure}[h!]
\centering
\includegraphics[width=0.95\textwidth]{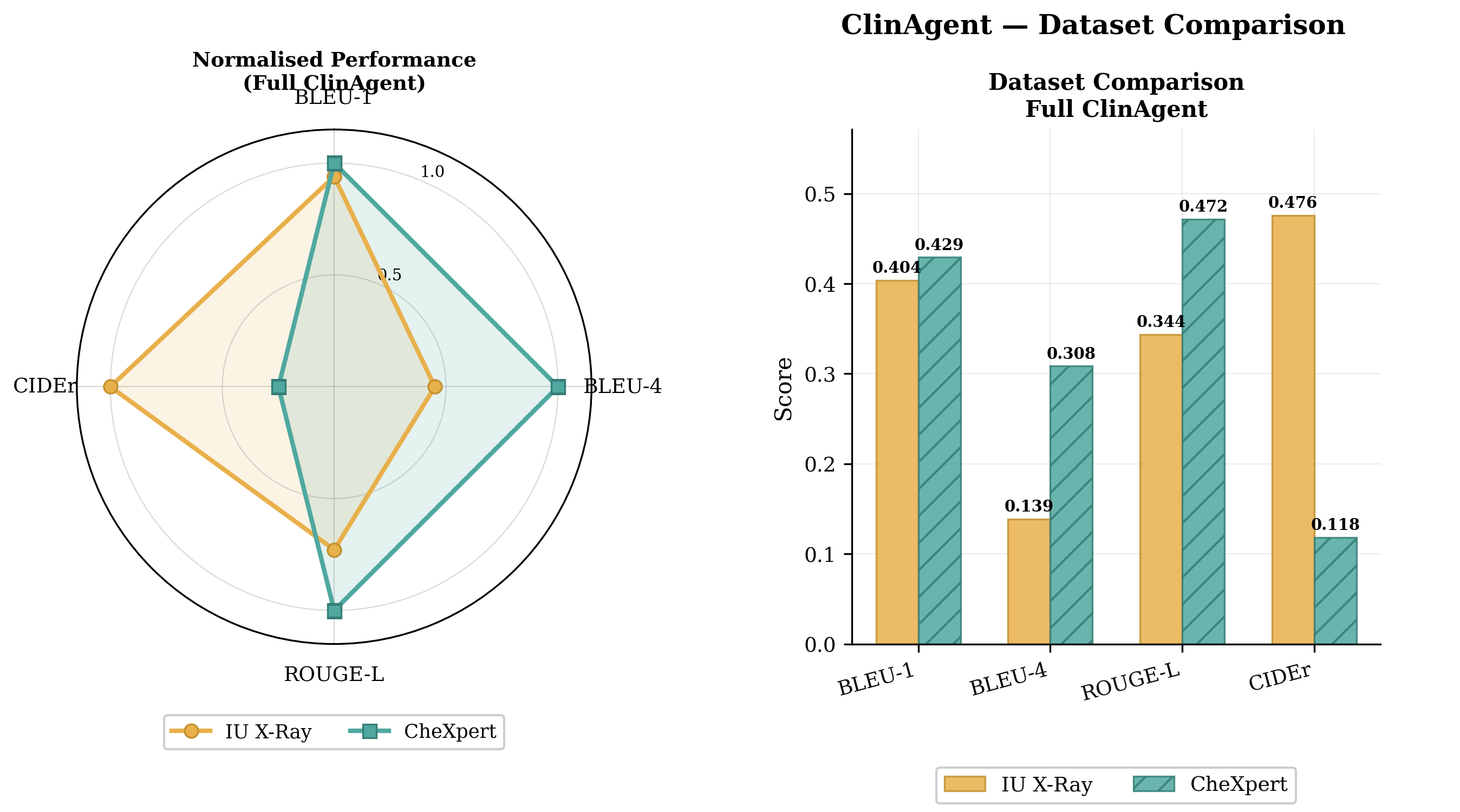}
\caption{Performance profiles of CogRad and its ablated variants across both datasets.}
\label{fig:dataset_comparison}
\end{figure}

\subsection{Clinical Accuracy and Hallucination Analysis}
\label{sec:hallucination}

The NLG metrics in Table~\ref{tab:main_results} measure surface agreement with a single reference report; they do not directly assess whether the pathologies a generated report states are actually present, nor how often it states pathologies that are not. Table~\ref{tab:clinical} reports RadGraph F1, CheXbert-14 and CheXbert-5 F1, and the hallucination/miss/entity-F1 statistics described in Section~\ref{sec:experiments} for both datasets.

\begin{table}[!htbp]
\caption{Clinical accuracy and hallucination metrics for CogRad on both datasets.}
\small
\label{tab:clinical}
\centering
\begin{tabular}{l cc}
\toprule
\textbf{Metric} & \textbf{IU X-Ray} & \textbf{CheXpert Plus} \\
\midrule
RadGraph F1 (partial)        & 0.316 & 0.162 \\
CheXbert-14 F1                & 0.287 & 0.330 \\
CheXbert-5 F1                 & 0.083 & 0.265 \\
Hallucination rate $\downarrow$ & 0.437 & 0.441 \\
Miss rate $\downarrow$        & 0.439 & 0.409 \\
Entity F1                     & 0.497 & 0.268 \\
\bottomrule
\end{tabular}
\end{table}

These results complicate the image suggested by Table~\ref{tab:main_results} alone. On IU X-Ray, CogRad achieves a higher RadGraph F1 (0.316) and a substantially lower hallucination rate (0.437) than the CheXpert Plus run, indicating that a non-trivial share of its generated content is entity-correct even where the exact phrasing diverges from the single available reference. CheXbert-5 F1 is low (0.083), reflecting that the five competition conditions are comparatively rare in IU X-Ray's outpatient population, so a small number of label disagreements has an outsized effect on this subset's F1.

On CheXpert Plus, the hallucination rate is 0.441 and is dominated by device- and finding-related terms: the most frequently hallucinated entities are pneumothorax, pleural effusion, lung opacity, consolidation, and the "no finding" category, while the most frequently missed entities are consolidation, effusion, lung opacity, lung lesion, and pneumothorax a pattern that points to the model both over-generating and under-localizing exactly the high-prevalence findings that dominate this dataset's case mix, rather than confusing them with rare or unrelated pathologies. Because CheXpert Plus is drawn substantially from inpatient and ICU populations, in which support devices and acute findings are reported far more frequently than in IU X-Ray's outpatient cohort, some portion of this gap reflects the underlying case mix rather than a uniform weakness of the model. The lower RadGraph F1 (0.162) on CheXpert Plus relative to IU X-Ray, despite CheXpert much stronger BLEU/ROUGE scores in Table~\ref{tab:main_results}, is itself informative: it shows directly that the disease-gated prefix strong n-gram performance on CheXpert Plus is not fully matched by entity-level factual accuracy, reinforcing why we treat the two metric families as complementary rather than substitutable, and why CogRad strong BLEU/ROUGE/CIDEr scores on CheXpert Plus (Section~\ref{sec:sota}) should not be read in isolation as a verdict on clinical quality.

\subsection{Clinical Explainability Analysis}

To assess the interpretability of the proposed framework, we visualize the cross-modal attention maps produced by the Verifier agent as Grad-CAM heatmaps. These visualizations highlight the image regions that contribute most strongly to the generation and verification of clinical findings, providing a transparent link between radiographic evidence and the generated narrative.

\begin{figure}[h!]
\centering
\includegraphics[width=0.99\textwidth]{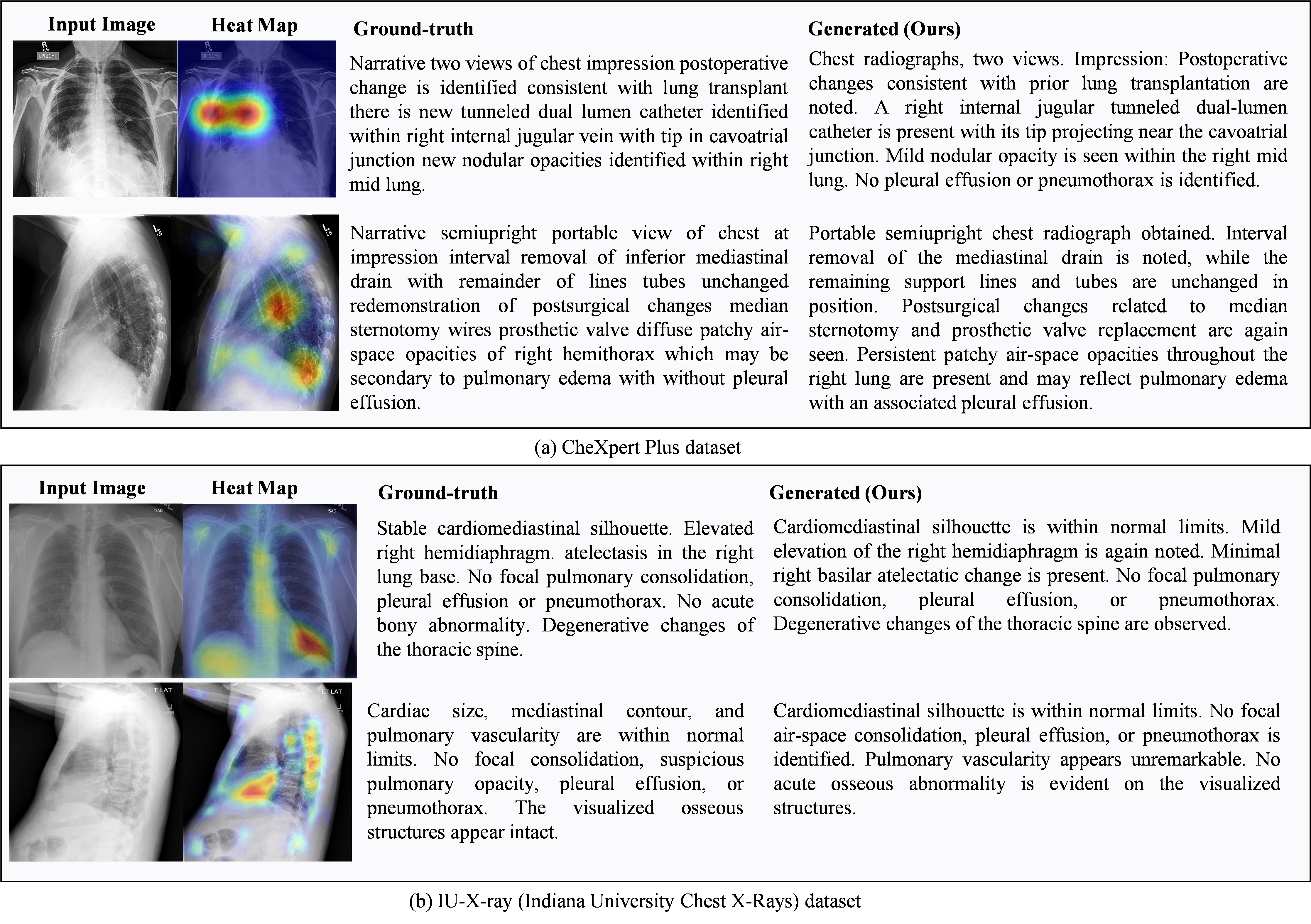}
\caption{Clinical explainability analysis for X-ray image diagnosis.}
\label{fig:XAM}
\end{figure}

Fig.~\ref{fig:XAM} presents representative examples comparing input chest radiographs, attention heatmaps, ground-truth reports, and CogRad's generated reports. On CheXpert Plus, the attention maps concentrate on the central thoracic region and right lung field, corresponding to postoperative changes, catheter placement, and pulmonary opacities described in both the reference and generated reports; the generated report captures the key clinical observations with strong semantic consistency with the ground truth. On IU X-Ray, the heatmaps concentrate on the lower thoracic and diaphragmatic regions, consistent with the reported right basilar atelectatic changes and elevated hemidiaphragm, and the generated report correctly reflects the absence of focal consolidation, pleural effusion, and pneumothorax while preserving the overall diagnostic interpretation. A lateral-view example similarly shows attention concentrated around the cardiomediastinal and lower lung regions, supporting findings related to normal cardiomediastinal contours, pulmonary vascularity, and the absence of acute thoracic abnormalities. Beyond these static attention maps, the sentence-level confidence head used by the inference-time re-examination loop (Section~\ref{sec:method}) offers a second, complementary form of explainability: rather than only indicating \emph{where} in the image a token attends, it provides a direct, per-sentence confidence score that can be surfaced to a clinician reviewing the report, flagging statements the model itself judged to be only weakly supported by the image before any human review takes place.

\subsection{Limitations and Future Directions}
\label{sec:limitations}

\subsubsection*{Limitations} Our evaluation surfaces three limitations worth stating plainly. First, although CogRad now outperforms every baseline on both datasets in NLG terms, its CIDEr score on CheXpert Plus (0.322) is still far below its RadGraph F1 advantage would suggest is achievable, indicating that strong n-gram overlap and strong entity-level factual accuracy do not automatically move together, and that NLG metrics alone should not be treated as a proxy for clinical accuracy. Second, the hallucination rate on CheXpert Plus (0.441) is slightly higher than on IU X-Ray (0.437) and is concentrated in high-prevalence findings and support devices; while this is consistent with CheXpert Plus's more acute, device-heavy case mix, we cannot yet fully separate the contribution of dataset distribution from a genuine weakness in how the Investigator allocates attention when multiple plausible findings co-occur.

\subsubsection*{Future directions} We see three natural extensions. Extending the entity lexicon used in our hallucination analysis to map specific sentences to specific anatomical regions, rather than scoring hallucination at the report level, would let the inference-time re-examination loop target re-generation more precisely instead of regenerating an entire report when only one sentence is flagged. Moreover, given the gap between CogRad's strong CheXpert Plus NLG scores and its more modest RadGraph F1 on the same dataset, directly incorporating a clinical-accuracy signal such as RadGraph or CheXbert into the training loss rather than using it only for evaluation is a promising direction for narrowing that gap.

\section{Discussion}
\label{sec:discussion}
The experimental results indicate that CogRad's structured multi-agent architecture brings a measurable benefit over single-pass generation baselines, though that benefit is concentrated rather than uniform across metrics and datasets. The clearest gains are on CheXpert Plus, where the availability of structured disease annotations lets the Scout agent learn an informative triage signal: CogRad reaches a BLEU-4 of 0.316 and a ROUGE-L of 0.480, well ahead of any baseline in the comparison, consistent with the idea that decomposing generation into triage, investigation, prefix construction, and verification captures clinically relevant structure that a single forward pass does not represent explicitly. The ablation study in Section~\ref{sec:ablation} suggests this benefit is concentrated in CIDEr rather than spread evenly across all NLG metrics on CheXpert Plus, with both the Scout's disease-classification loss and the Verifier's entailment loss contributing a small, consistently positive effect on that metric specifically. The results on IU X-Ray reinforce the same pattern: CogRad reaches a BLEU-4 of 0.201 and a CIDEr of 0.724, both ahead of every baseline in the comparison. Even so, CogRad's own RadGraph F1 on IU X-Ray (0.316) is higher than its RadGraph F1 on CheXpert Plus (0.162), despite CheXpert Plus being where its BLEU and ROUGE-L scores are strongest a reminder that leading on n-gram-based metrics on a given dataset does not guarantee a proportionally larger lead in entity-level clinical accuracy, and the reason we report RadGraph F1~\cite{jain2021radgraph} and CheXbert F1~\cite{smit2020combining} alongside the standard NLG metrics rather than treating either family as sufficient on its own. More broadly, CogRad's design choice to maintain a continuous tensor-level visual representation across all four agents, rather than collapsing to text at any intermediate stage, gives the Verifier's cross-modal attention weights a natural interpretability benefit: they double as token-level grounding maps without any auxiliary explainability module or post-hoc analysis, letting a reviewing radiologist see which image regions a given statement is anchored to. This is a meaningfully different starting point from approaches that treat visual encoding and language generation as loosely coupled stages connected only through text.
 
\section{Conclusion}
\label{sec:conclusion}
 
We presented CogRad, a cognitively-inspired multi-agent framework for chest X-ray report generation that structures the generation process around four stages of a radiologist's reading workflow: global triage over learned anatomical regions, focused investigation of suspicious areas, disease-gated visual prompting of a language model, and a Verifier agent that supervises training with a visual-entailment loss and, at inference, re-examines and selectively regenerates low-confidence sentences in its own draft. Across CheXpert Plus and IU X-Ray, CogRad's gains are real but uneven: it outperforms recent baselines on standard NLG metrics on both CheXpert Plus and IU X-Ray, while its RadGraph F1 and CheXbert F1 scores show that this NLG advantage does not translate into a proportionally larger advantage in entity-level clinical accuracy. We see this unevenness, together with the open ablation question on IU X-Ray, as the most useful outcome of this work alongside the architecture itself: it indicates concretely where a structured, self-checking agent pipeline helps, where current overlap-based metrics fail to reflect that help, and where further evaluation is still needed before the gains reported here should be treated as settled.
 

\section*{Declarations}
\begin{itemize}
\item \textbf{Ethics approval and consent to participate:} Not applicable.
\item \textbf{Funding:} No funding.
\item \textbf{Declaration of competing interest:} The authors declare that they have no known competing financial interests or personal relationships that could have appeared to influence the work reported in this paper.
\item \textbf{Consent for publication:} Not applicable.
\item \textbf{Data availability:} Available from corresponding author upon request.
\item \textbf{CRediT authorship contribution statement:} 
Saif Ur Rehman Khan \& Hasaan Maqsood: Conceptualization, Data curation, Methodology, Software, Validation, Writing original draft \& Formal analysis. Muhammad Nabeel Asim, Sebastian Vollmer \& Andreas Dengel: Conceptualization, Funding acquisition, Review.
\end{itemize}
 
\section*{Abbreviations}\label{sec:abbrev}
Table~\ref{tab:abbreviations} lists the key abbreviations
used throughout this paper.

\begin{table}[!htbp]
\centering
\caption{Abbreviations and Definitions}
\label{tab:abbreviations}
\begin{tabular}{lp{0.7\linewidth}}
\toprule
\textbf{Abbreviation} & \textbf{Definition} \\
\midrule
\multicolumn{2}{l}{\textbf{Clinical \& Medical Imaging}} \\
\midrule
RRG   & Radiology Report Generation \\
CXR   & Chest X-Ray \\
AI    & Artificial Intelligence \\
XAI   & Explainable Artificial Intelligence \\
\midrule
\multicolumn{2}{l}{\textbf{Model Components \& Architecture}} \\
\midrule
LLM   & Large Language Model \\
VLM   & Vision--Language Model \\
MHA   & Multi-Head Attention \\
MLP   & Multi-Layer Perceptron \\
LN    & Layer Normalisation \\
LoRA  & Low-Rank Adaptation \\
\midrule
\multicolumn{2}{l}{\textbf{Training \& Optimisation}} \\
\midrule
BCE   & Binary Cross-Entropy Loss \\
MSE   & Mean Squared Error \\
CE    & Cross-Entropy Loss \\
AdamW & Adam Optimiser with Decoupled Weight Decay \\
AMP   & Automatic Mixed Precision \\
bf16  & Brain Float 16-bit Precision \\
\midrule
\multicolumn{2}{l}{\textbf{Evaluation Metrics}} \\
\midrule
BLEU  & Bilingual Evaluation Understudy \\
ROUGE & Recall-Oriented Understudy for Gisting Evaluation \\
CIDEr & Consensus-based Image Description Evaluation \\
F1    & F1 Score (Harmonic Mean of Precision and Recall) \\
NLG   & Natural Language Generation \\
\midrule
\multicolumn{2}{l}{\textbf{Datasets}} \\
\midrule
IU X-Ray    & Indiana University Chest X-Ray Dataset \\
CheXpert    & Chest Expert -- Stanford Large-Scale Chest Radiograph Dataset \\
CheXpert Plus & Extended CheXpert Dataset with Radiology Reports and 14 Pathology Labels \\
\bottomrule
\end{tabular}
\end{table}
\newpage
\bibliography{sn-bibliography}

\end{document}